\documentclass[10pt,twocolumn,letterpaper]{article}

\usepackage[pagenumbers]{cvpr} 

\usepackage{graphicx}
\usepackage{amsmath}
\usepackage{amssymb}
\usepackage{booktabs}

\usepackage{subfiles}
\usepackage{multirow}
\usepackage{verbatim}
\usepackage{pifont}%
\newcommand{\xmark}{\ding{53}}%
\usepackage[pagebackref,breaklinks,colorlinks]{hyperref}

\usepackage[capitalize]{cleveref}
\crefname{section}{Sec.}{Secs.}
\Crefname{section}{Section}{Sections}
\Crefname{table}{Table}{Tables}
\crefname{table}{Tab.}{Tabs.}

\def\cvprPaperID{6063} 
\def\confName{CVPR}
\def\confYear{2022}

\begin{document}


\title{ADAPT: Vision-Language Navigation with Modality-Aligned Action Prompts}

\author{Bingqian Lin\textsuperscript{1}\thanks{Part of this work was done during an internship in Huawei Noah's Ark Lab.}, Yi Zhu\textsuperscript{2}, Zicong Chen\textsuperscript{1}, Xiwen Liang\textsuperscript{1}, Jianzhuang Liu\textsuperscript{2}, Xiaodan Liang\textsuperscript{1}\thanks{Corresponding Author}\\
\textsuperscript{1}Shenzhen Campus of Sun Yat-sen University \textsuperscript{2}Huawei Noah's Ark Lab\\
{\tt\small  \{linbq6@mail2,chenzc7@mail2,liangxw29@mail2, liangxd9@mail\}sysu.edu.cn} \\ {\tt\small  \{zhuyi36,   liu.jianzhuang\}@huawei.com}}

\maketitle

\begin{abstract}
Vision-Language Navigation (VLN) is a challenging task that requires an embodied agent to perform action-level modality alignment, i.e., make instruction-asked actions sequentially in complex visual environments. 
Most existing VLN agents learn the instruction-path data directly and cannot sufficiently explore action-level alignment knowledge inside the multi-modal inputs.
In this paper, we propose  mod\textbf{A}lity-aligne\textbf{D} \textbf{A}ction \textbf{P}romp\textbf{T}s (ADAPT), which provides the VLN agent with action prompts to enable the explicit learning of action-level modality alignment to pursue successful navigation. Specifically, an action prompt is defined as a modality-aligned pair of an image sub-prompt and a text sub-prompt, where the former is a single-view observation and the latter is a phrase like ``walk past the chair''. When starting navigation, the instruction-related action prompt set is retrieved from a pre-built action prompt base and passed through a prompt encoder to obtain the prompt feature. Then the prompt feature is concatenated with the original instruction feature and fed to a multi-layer transformer for action prediction. To collect high-quality action prompts into the prompt base, we use the Contrastive Language-Image Pretraining (CLIP) model which has powerful cross-modality alignment ability. A modality alignment loss and a sequential consistency loss are further introduced to enhance the alignment of the action prompt and enforce the agent to focus on the related prompt sequentially. Experimental results on both R2R and RxR show the superiority of ADAPT over state-of-the-art methods. 

\end{abstract}

\section{Introduction}
\begin{figure}[t]
\begin{centering}
\includegraphics[width=0.95\linewidth]{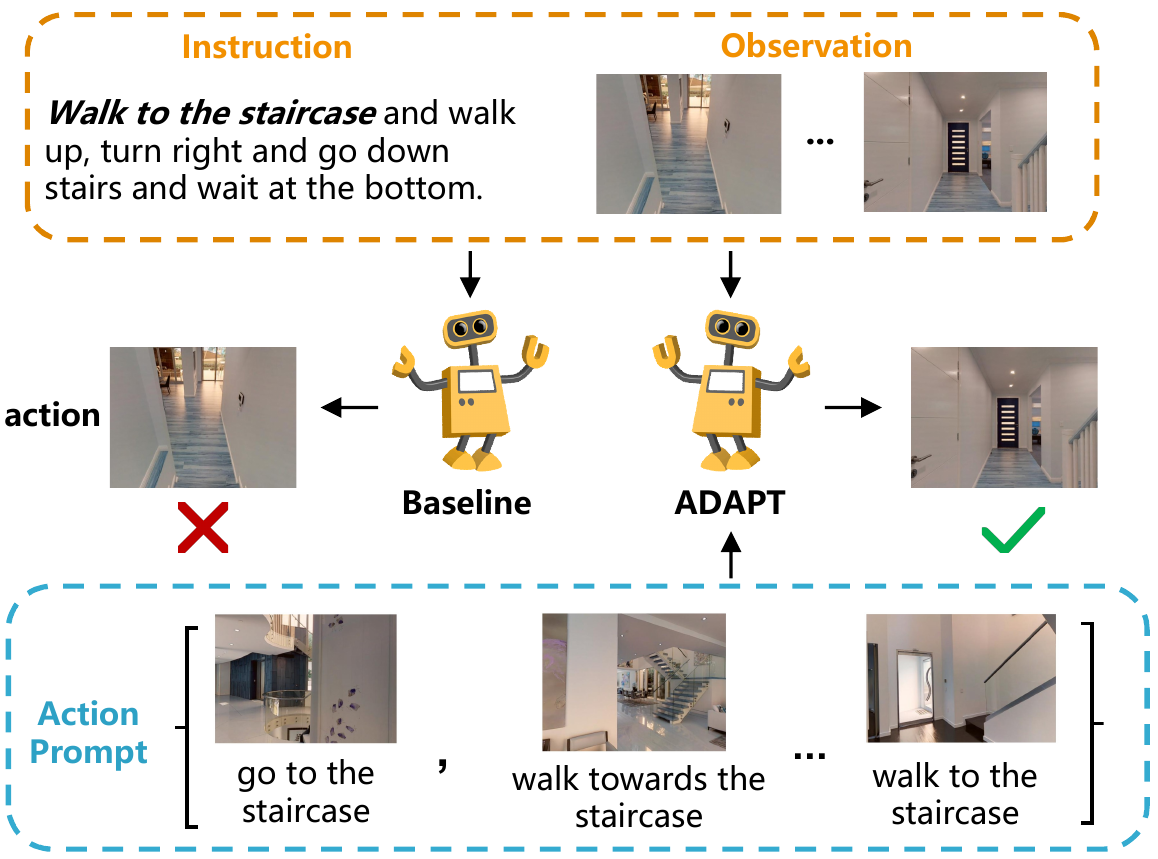}
\par\end{centering}
\caption{\label{fig:motivation}The action decision comparison between a baseline \cite{hong2021vln} and our ADAPT. With the help of action prompts related to ``walk to the staircase'' in the instruction, our ADAPT successfully makes correct action from the current observation.
}
\vspace{-0.6cm}
\end{figure}

In the Vision-Language Navigation (VLN)  task \cite{anderson2018vision, chen2019touchdown}, an embodied agent is required to navigate through complex scenes following a given language instruction. To accomplish successful navigation, the agent needs to implement both object-level and action-level modality alignment accurately given the instruction and visual observations. For example, given an instruction of ``exit the bedroom'', the agent should not only locate the ``bedroom'' in its observation but also find the door of the bedroom to make the action of ``exit''. With great potential in the applications such as in-home robots and personal assistants, VLN has received wide spread attention in the robotic visual applications.  

Early VLN approaches explore diverse data augmentation strategies \cite{tan2019learning, fried2018speaker, liu2021vision, fu2020counterfactual}, efficient learning paradigms \cite{wang2019reinforced, zhu2020vision,huang2019transferable, li2019robust,Zhu2020BabyWalkGF} and useful model architecture \cite{wang2019reinforced, ma2019self, deng2020evolving, hong2020language} to improve the agent  performance. Motivated by the significant progress made by large-scale cross-modal pre-trained models in vision-language tasks \cite{su2020vl,li2019visualbert,chen2020uniter,li2020unicoder,li2020oscar}, more and more works attempt to introduce the pretraining paradigms and models into the VLN task. PREVALENT \cite{hao2020towards} pretrains the model on a large amount of image-text-action triplets in a self-supervised learning manner.   VLN$\circlearrowright$BERT \cite{hong2021vln} introduces a recurrent function into the pretrained models to make the VLN agent time-aware. 
Although the object-level alignment ability may be significantly enhanced through the pretraining process, these VLN agents still learn the action-level modality alignment in an implicit way, which largely limits the robust action decision under different scenes.

Recently, the prompt engineering paradigm has shown great potential in endowing pretrained models with diverse capabilities through simply providing prompts designed by experts or optimized with task-specific objectives \cite{liu2021pre,lester2021the,zhong2021factual,zhou2021learning,tsimpoukelli2021multimodal}. 
Inspired by this, we propose to introduce the prompt into the VLN task to improve the action-level modality alignment ability of the pretrained VLN agents. 
To this end, we propose mod\textbf{A}lity-aligne\textbf{D} \textbf{A}ction \textbf{P}romp\textbf{T}s (ADAPT), where the agent is provided with explicit {\it action prompts} to make action decision. 
An action prompt contains a pair of multi-modal sub-prompts, where the image sub-prompt is a single-view observation indicating a salient visual object or location, and the paired text sub-prompt is an object-related action phrase like ``go to the staircase''. 

Before navigating, the instruction-related action prompts are retrieved from a pre-constructed action prompt base. 
Then the action prompts are passed through a prompt encoder and the output feature is concatenated with the original instruction feature. 
The prompt-based instruction feature, together with the visual feature are fed to a multi-layer transformer for making action decision.
Note that different from the common prompt engineering methods which change the output prediction form of a downstream task by introducing the prompt \cite{liu2021pre}, in this work, we keep the same form of the action prediction as the baseline model and focus on the design of the prompts.
Through these provided action prompts, the agent can learn the action-level modality alignment explicitly and make robust actions in different scenes. To enhance the discriminative power of the action prompts and enforce the agent to attend to related action prompts at each timestep, a modality alignment loss and a sequential consistency loss are further introduced into the training. Fig.~\ref{fig:motivation} presents an action decision comparison between the baseline agent \cite{hong2021vln} and our ADAPT. 
As shown in Fig.~\ref{fig:motivation}, with the help of the action prompts related to ``walk to the staircase'', our ADAPT can choose the correct action in the given observations to navigate successfully.

To collect high-quality action prompts into the action prompt base, we resort to the recently developed Contrastive Language-Image Pretraining (CLIP) \cite{radford2021learning} model which has powerful cross-modal object/location-level alignment ability. Concretely, the image sub-prompt is obtained by  retrieving object/location-related images using CLIP from the action image sequence where each image contains the action information itself. 
The text sub-prompt is derived through a simple nearest-verb-search scheme.

Experimental results on both Room-to-Room (R2R) \cite{anderson2018vision} and Room-across-Room (RxR) \cite{ku2020room} benchmarks show the superiority of our proposed ADAPT over the state-of-the-art methods, demonstrating that introducing explicit action prompts is promising for improving the agent navigation performance. Our ablation study indicates the effectiveness of each method component and the good generalization ability of ADAPT. 
The visualization analysis also shows its good interpretability. 

To summarize, the main contributions of this paper are: 1) We propose modality-aligned action prompts (ADAPT) to enforce the VLN agent to learn cross-modal action knowledge explicitly for improving action decision during navigation. To the best of our knowledge, this is the first attempt to develop prompt-based agents in the VLN task. 2) We develop a modality alignment loss and a sequential consistency loss for enabling efficient learning of action prompts. The  Contrastive Language-Image Pretraining (CLIP) model is employed to ensure the quality of the action prompts. 3) ADAPT establishes new state-of-the-art results on both R2R and RxR. It also shows good interpretability and generalization ability. 

\section{Related Work}
\textbf{Vision-Language Navigation.}
Given the language instruction, a VLN agent is required to follow it to reach a predefined goal position. Early methods usually employ a sequence-to-sequence model architecture \cite{fried2018speaker,tan2019learning,zhu2020vision}. Speaker-follower  \cite{fried2018speaker} introduces  synthetic instructions to alleviate the annotation burden of instructions. EnvDrop \cite{tan2019learning} develops an environmental dropout strategy to generate augmented data by mimicking unseen environments.  

Recently, large-scale vision-language pretraining models \cite{su2020vl,li2019visualbert,chen2020uniter,li2020unicoder,li2020oscar} have shown significant superiority on multiple vision-language understanding tasks like Visual Commonsense Reasoning  \cite{zellers2019from} and Visual Question Answering  \cite{antol2015vqa}. Inspired by this, more and more works have introduced vision-language pretrained models into the VLN area \cite{hao2020towards,hong2021vln,qi2021the}.  PREVALENT \cite{hao2020towards} collects plenty of  image-text-action triplets to pretrain the agent with  self-supervised tasks such as attended masked language modeling and action prediction. VLN$\circlearrowright$BERT \cite{hong2021vln} adds a recurrent function to help the agent recognize time-dependent input. 
However, in these pretrained VLN methods, the agent learns the relationship between the action decision and multi-modal information implicitly, leading to inefficient training and limited generalization abilities. 
In this paper, we take the first step to develop a prompt-based VLN agent, which receives explicit action prompts indicating cross-modal action knowledge for assisting the action decision during navigation.

\begin{figure*}[t]
\begin{centering}
\includegraphics[width=0.95\linewidth]{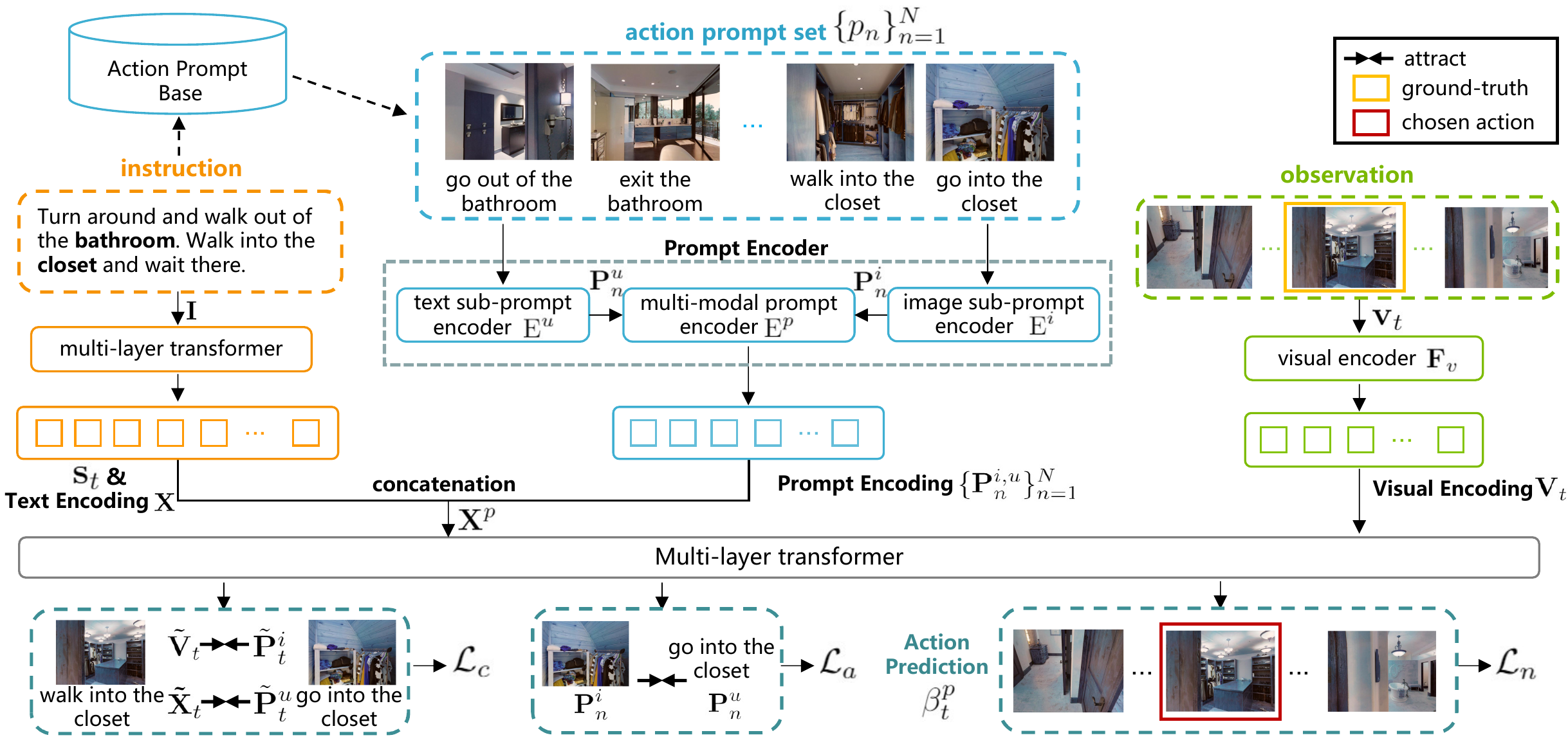}
\par\end{centering}
\caption{\label{fig:overview}Overview of our ADAPT. At timestep $t$, the agent receives the instruction, visual observation, and retrieved action prompts. The action prompts are passed through the prompt encoder and the output feature is concatenated with the instruction encoding $\mathbf{X}$ to obtain prompt-based instruction feature $\mathbf{X}^{p}$. The action decision is made based on $\mathbf{X}^{p}$ and the visual encoding $\mathbf{V}_{t}$. The navigation loss $\mathcal{L}_{n}$, the sequential consistency loss $\mathcal{L}_{c}$ and the modality alignment loss $\mathcal{L}_{a}$ are applied to optimize ADAPT. (Best viewed in color.)
}
\vspace{-0.4cm}
\end{figure*}

\textbf{Prompt Engineering.}
Recent studies have shown that prompts play a vital role in improving pretrained language models in many downstream NLP tasks \cite{liu2021pre,lester2021the,zhong2021factual,li2021prefix,brown2020language,raffel2020exploring}. 
Jiang et al. \cite{jiang2020how} apply the text mining and paraphrasing techniques to generate the candidate prompts and choose the one with the highest accuracy. For facilitating prompt learning, Shin et al. \cite{shin2020autoprompt} propose to generate prompts automatically through the gradient-based search. Lately, some works \cite{lester2021the,zhong2021factual,li2021prefix} propose to generate continuous prompts instead of hand-crafted text prompts. 

Inspired by the progress that prompt learning has made in NLP, some works attempt to introduce it into the pretrained vision-language models  recently \cite{zhou2021learning,yao2021cpt,tsimpoukelli2021multimodal}. CoOp \cite{zhou2021learning} models the context in prompts using continuous representations and keeps the pretrained model parameters fixed to conduct end-to-end learning. 
CPT \cite{yao2021cpt} reformulates  the visual grounding task into a fill-in-the-blank problem with color-based cross-modal prompts.
Frozen \cite{tsimpoukelli2021multimodal} encodes the image as a sequence of continuous embeddings to serve as the prefix to implement multi-modal few-shot learning. 
In the light of the prompt engineering paradigm, we introduce the modality-aligned action prompts during navigation for enabling VLN agents to learn cross-modal action knowledge explicitly. Through these action prompts, the agent can effectively learn action-level modality alignment for implementing successful navigation. 

\textbf{Contrastive Language-Image Pretraining (CLIP).}
CLIP \cite{radford2021learning} is a  large-scale pre-trained model that relies on natural language supervision to learn visual representations. For an image-text pair, a visual encoder and a text encoder are used to encode the input representations independently. And the dot product between the two encoder's output serves as the alignment score of the image-text pair. 
Through training on 400M noisy image-text pairs, CLIP has shown strong zero-shot capabilities on benchmarks such as ImageNet classification. 
Recently, some works propose to resort to the knowledge learned in CLIP to improve the generalization ability of downstream models, including object detection \cite{Gu2021open}, image manipulation \cite{patashnik2021styleclip}, and vision-language tasks \cite{shen2021how}.
In this paper, we employ CLIP to retrieve the image containing instruction-referred visual object/location in a specific action image sequence for building action prompts. With the powerful cross-modal alignment ability of CLIP, the  instruction-referred visual object/location images can be effectively retrieved for ensuring the quality of the action prompts.


\section{Method}
The overview of our ADAPT is given in Fig.~\ref{fig:overview}. Before navigation, the agent retrieves the instruction-related action prompts from the action prompt base. Then the agent makes the action decision at each timestep based on the given  instruction, the visual observation, and retrieved action prompts. The navigation is optimized by the navigation loss $\mathcal{L}_{n}$, the sequential consistency loss $\mathcal{L}_{c}$, and the modality alignment loss $\mathcal{L}_{a}$. 

\subsection{VLN Problem Setup}
\label{sec: VLN Problem Setup}
Given a language instruction $\mathbf{I}=\{w_{0},...,w_{L}\}$ with $L$ words, a VLN agent is required to find a route from a start viewpoint $c_{0}$ to the target viewpoint $c_{T}$.  At each timestep $t$, the agent  observes a panoramic view, which contains 36 image views  $\{o_{t,i}\}_{i=1}^{36}$. Each image view $o_{t,i}$ includes an RGB image $b_{t,i}$ accompanied with its orientation ($\theta_{t,i}^{1}$,$\theta_{t,i}^{2}$), where $\theta_{t,i}^{1}$ and $\theta_{t,i}^{2}$ are the
angles of heading and elevation, respectively.  With the instructions and current visual observations, the agent infers the action for each step $t$ from the candidate actions list, which consists of $J$ neighbors of the current node in the navigation connectivity graph $\mathcal{G}=(V,E)$  and a stop action. $V$ and $E$ represent the nodes and edges in the navigation connectivity graph, respectively. 

\subsection{VLN Agent with Action Prompts}
\label{sec:VLN agent with Action Prompt}

\subsubsection{Baseline Agent}
Our baseline agent follows the architecture of VLN$\circlearrowright$BERT \cite{hong2021vln}, which is a multi-layer transformer model consisting of the self-attention module and cross-modal attention module. At each timestep, the model receives the cross-modal inputs for the action prediction.

\textbf{Visual Input.}
For each image view $o_{t,i}$ in the candidate views at timestep $t$, a pretrained Convolutional Neural Network (CNN) \cite{hong2021vln} or a transformer \cite{shen2021how} is applied in advance to extract image feature $\mathbf{v}_{t,i}$. Then $\mathbf{v}_{t,i}$ is projected by a visual encoder $\mathbf{F}_{v}$ \cite{hong2021vln} to get the visual encoding $\mathbf{V}_{t,i}$: 
\vspace{-0.1cm}
\begin{equation}
\mathbf{V}_{t,i} = \mathbf{F}_{v}(\mathbf{v}_{t,i};\theta_{v}),   
\end{equation}
where $\theta_{v}$ denotes the parameters of $\mathbf{F}_{v}$. The set $\mathbf{V}_{t}=\{\mathbf{V}_{t,i}\}_{i=1}^{36}$ denotes the candidate visual encodings at timestep $t$.

\textbf{Language Input.}
When initialization, the instruction encoding $\mathbf{X}$ and the initialized state feature $\mathbf{s}_{0}$ are obtained by feeding the instruction sequence $\mathbf{I}$ together with [CLS] and [SEP] tokens to the self-attention module in the transformer:
\vspace{-0.2cm}
\begin{equation}
\mathbf{s}_{0}, \mathbf{X} = \mathrm{SelfAttn}(\mathrm{Concat}([\mathrm{CLS}], \mathbf{I}, [\mathrm{SEP}]);\theta_{s}^{1}),
\end{equation}
where $\mathrm{Concat}(\cdot)$ represents the concatenation operation, and $\theta_{s}^{1}$ denotes the parameters of the self-attention module.  $\mathbf{s}_{0}$  will be updated to obtain $\mathbf{s}_{t}$ at each timestep $t$.

\textbf{Action Decision.}
During the action decision at timestep $t$, the state feature $\mathbf{s}_{t}$ is concatenated with the visual feature $\mathbf{V}_{t}$ to obtain the state-visual feature $\mathbf{K}_{t}$. 
Then the cross-modal attention $\mathbf{\alpha}_{t}$ between $\mathbf{K}_{t}$ and the instruction feature $\mathbf{X}$ is calculated to update $\mathbf{K}_{t}$:
\begin{equation}
\mathbf{\tilde{K}}_{t}, \mathbf{\alpha}_{t}= \mathrm{CrossAttn}(\mathbf{K}_{t}, \mathbf{X};\theta_{c}),     
\end{equation}
where $\theta_{c}$ represents the parameters of the cross-modal attention module. The attended instruction feature $\mathbf{\tilde{X}}_{t}$ is derived by weighting the instruction feature $\mathbf{X}$ by $\mathbf{\alpha}_{t}$. The updated state-visual feature $\mathbf{\tilde{K}}_{t}$ is further fed to another self-attention module $\mathrm{SelfAttn}(\cdot)$ to obtain the attention scores $\mathbf{\beta}_{t}$ of the state feature $\mathbf{s}_{t}$ over the visual feature $\mathbf{V}_{t}$, which is also  treated as the action prediction probability:
\begin{align}
\label{eq:self attention}
\mathbf{\beta}_{t} &= \mathrm{SelfAttn}(\mathbf{\tilde{K}}_{t};\theta_{s}^{2}),
\end{align}
where $\theta_{s}^{2}$ represents the module parameters. The attended visual feature $\mathbf{\tilde{V}}_{t}$ is obtained through weighting the visual feature $\mathbf{V}_{t}$ by $\mathbf{\beta}_{t}$. Then $\mathbf{\tilde{X}}_{t}$ and $\mathbf{\tilde{V}}_{t}$ are used for updating the state feature $\mathbf{s}_{t}$ which is used for the next timestep action prediction. 
For more model details, refer to \cite{hong2021vln}.
\begin{figure}[t]
\begin{centering}
\includegraphics[width=0.95\linewidth]{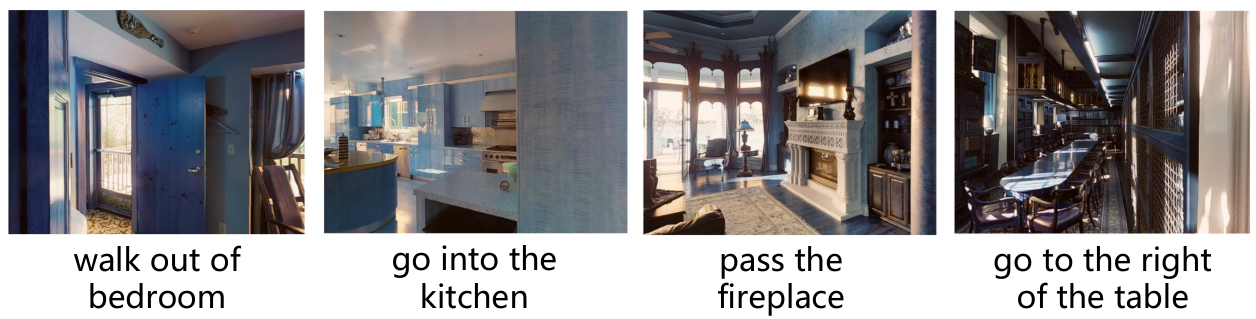}
\par\end{centering}
\caption{\label{fig:prompt}Examples of action prompts. 
}
\vspace{-0.6cm}

\end{figure}

\begin{figure*}[t]
\begin{centering}
\includegraphics[width=0.95\linewidth]{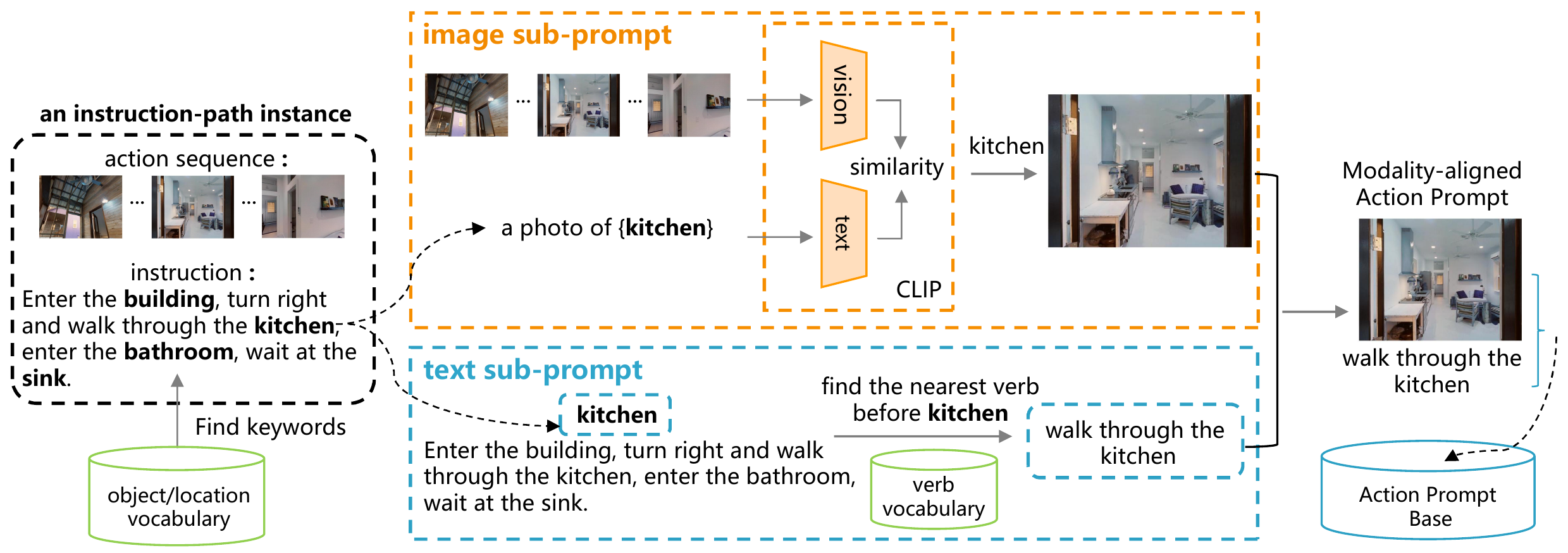}
\par\end{centering}
\caption{\label{fig:prompt base construction} Illustration of action prompt collection for building the action prompt base. Given a training instruction-path instance, the image and text sub-prompts are firstly obtained via CLIP and nearest verb search, respectively. Then the multi-modal sub-prompts related to the same visual object/location and action are aligned to form an action prompt. Here the word ``kitchen'' is taken as an example.
}
\vspace{-0.6cm}
\end{figure*}

\subsubsection{Action Prompts}
Before describing our prompt-based VLN agent, we first define the action prompts.
An action prompt is a modality-aligned pair of an image sub-prompt and a text sub-prompt, where the former is a single-view observation and the latter is an action phrase. The observation indicates a salient visual object or a location. The action phrase contains two main elements, i.e., a word/phrase representing the {\it action} such as ``exit'' or ``walk into'', and a {\it  object/location} word such as ``chair'' or ``bedroom''. Fig.~\ref{fig:prompt} shows some examples of the action prompts. From Fig.~\ref{fig:prompt} we can find that an action prompt not only contains an aligned visual object or location in both modalities but also indicates the modality-aligned action knowledge. For example, the paired image sub-prompt of the text sub-prompt ``walk out of bedroom'' contains the appearance of the bedroom and its door, through which the agent can complete the action of ``walk out of'' the bedroom. Therefore, by explicitly providing the action prompts into the training, the agent is able to better explore the cross-modal action knowledge which is important for guiding correct action decision. The construction of the action prompt base is described in Sec.~\ref{sec:Action Prompt Base Construction}.

\subsubsection{Action Decision with Action Prompts}
At the beginning of the navigation, the agent retrieves instruction-correlated action prompts from the action prompt base. Specifically, the object/location-related action phrases in the given instruction are derived following the strategy for obtaining text sub-prompts (see Sec.~\ref{sec:Action Prompt Base Construction}). Then the sentence similarity between each object/location-related action phrase and the text sub-prompts in the prompt base is calculated to retrieve the instruction-related action prompt set $\{p_{n}\}_{n=1}^{N}$, where $N$ is the size of the set.  

With $\{p_{n}\}_{n=1}^{N}$, we obtain the prompt encoding $\{\mathbf{P}_{n}^{i,u}\}_{n=1}^{N}$ through the prompt encoder (see Fig.~\ref{fig:overview}). The prompt encoder consists of two single-modal sub-prompt encoders and a multi-modal prompt encoder. Denote the image and text sub-prompts in the action prompt $p_{n}$ as $p_{n}^{i}$ and $p_{n}^{u}$, respectively, i.e., $p_{n}=\{p_{n}^{i}, p_{n}^{u}\}$. $p_{n}^{i}$ and $p_{n}^{u}$ are firstly passed through the single-modal sub-prompt encoders to obtain the sub-prompt features $\mathbf{P}_{n}^{i}$ and $\mathbf{P}_{n}^{u}$:
\begin{align}
\mathbf{P}_{n}^{i} &= \mathrm{E}^{i}(p_{n}^{i};\theta^{i}),\\
\mathbf{P}_{n}^{u} &= \mathrm{E}^{u}(p_{n}^{u};\theta^{u}),
\end{align}
where $\mathrm{E}^{i}(\cdot)$ with parameters $\theta^{i}$ and $\mathrm{E}^{u}(\cdot)$ with parameters $\theta^{u}$ represent the image sub-prompt encoder and text sub-prompt encoder, respectively.  Then $\mathbf{P}_{n}^{i}$ and $\mathbf{P}_{n}^{u}$ are fed to the multi-modal prompt encoder $\mathrm{E}^{p}(\cdot)$ to obtain the prompt encoding $\mathbf{P}_{n}^{i,u}$:
\begin{equation}
\mathbf{P}_{n}^{i,u} = \mathrm{E}^{p}(\mathrm{Concat}(\mathbf{P}_{n}^{i},\mathbf{P}_{n}^{u});\theta^{p}), 
\end{equation}
where $\theta^{p}$ denotes the parameters of  $\mathrm{E}^{p}(\cdot)$, and $\mathrm{Concat}(\cdot)$ is the concatenation operation. In our ADAPT, the encoders $\mathrm{E}^{i}(\cdot)$,  $\mathrm{E}^{u}(\cdot)$ and  $\mathrm{E}^{p}(\cdot)$ consists of one linear layer followed by the dropout operation to reduce the over-fitting.

With the prompt encoding $\{\mathbf{P}_{n}^{i,u}\}$ and the instruction encoding $\mathbf{X}$, we obtain the prompt-based instruction feature $\mathbf{X}^{p}$ by simply concatenating $\mathbf{X}$ and $\{\mathbf{P}_{n}^{i,u}\}$.
Then the state-visual feature $\mathbf{K}_{t}$ is updated based on the cross-modal attention $\mathbf{\alpha}_{t}^{p}$ between $\mathbf{K}_{t}$ and $\mathbf{X}^{p}$: 
\begin{equation}
\mathbf{\tilde{K}}_{t}^{p}, \mathbf{\alpha}_{t}^{p} = \mathrm{CrossAttn}(\mathbf{K}_{t}, \mathbf{X}^{p};\theta_{c}).  
\end{equation}
$\mathbf{\alpha}_{t}^{p}$ is then split to $\mathbf{\alpha}_{t}^{p_{1}}$ and $\mathbf{\alpha}_{t}^{p_{2}}$ for obtaining different attended features. Concretely, the attended instruction feature $\mathbf{\tilde{X}}_{t}$ is derived via weighting  $\mathbf{X}$ by $\mathbf{\alpha}_{t}^{p_{1}}$. The attended image sub-prompt feature $\mathbf{\tilde{P}}_{t}^{i}$ and the attended text sub-prompt feature $\mathbf{\tilde{P}}_{t}^{u}$ are obtained through weighting $\mathbf{P}_{n}^{i}$ and $\mathbf{P}_{n}^{u}$ by $\mathbf{\alpha}_{t}^{p_{2}}$. $\mathbf{\tilde{P}}_{t}^{i}$ and $\mathbf{\tilde{P}}_{t}^{u}$ are used for calculating the sequential consistency loss $\mathcal{L}_{c}$. $\mathbf{\tilde{X}}_{t}$ is used for updating the state feature like the baseline agent.  Finally, the prompt-based action prediction probability $\mathbf{\beta}_{t}^{p}$ is obtained by feeding $\mathbf{\tilde{K}}_{t}^{p}$ into the self-attention module like that in Eq.~\ref{eq:self attention}. 

\subsection{Construction of the Action Prompt Base}
\label{sec:Action Prompt Base Construction}
Although it is easy to assign an object category label to an image through object recognition, associating an image with an action phrase is not straightforward. To better align the image and the action phrase to form the action prompt, we design a two-branch scheme to collect the image and text sub-prompts, as shown in Fig.~\ref{fig:prompt base construction}. 
At first, for an instruction-path instance in the training dataset, we use a pre-constructed visual object/location vocabulary to find the referred visual objects/locations in the instruction. Then for each visual object/location, we obtain the related image and text sub-prompts separately as described below.

Note that the ground-truth path sequence contains a set of single-view images, each of which indicates an  action needed to make at the specific timestep. Therefore, for deriving the image sub-prompt in an action prompt, we only need to retrieve the object/location-related image from the ground-truth path sequence, which itself contains the action information. 
Instead of resorting to existing object classifiers or detectors trained on a fixed set of class categories \cite{he2016deep, ren2017faster}, we use  CLIP \cite{radford2021learning} which shows  excellent zero-shot cross-modal alignment ability to locate the object/location-related image. 
To adapt to the inference process of CLIP, we replace the \{CLASS\} token in the phrase ``a photo of \{CLASS\}'' with the visual object/location whose category label is $c$. The probability that an image $\boldsymbol{B}$  in the action sequence belongs to the class $c$ is calculated by:
\begin{equation}
p(y=c|\boldsymbol{B}) = \frac{\mathrm{exp}(\mathrm{sim}(\boldsymbol{b},\boldsymbol{w}_{c})/\tau_{1})}{\sum_{i=1}^{M}(\mathrm{exp}(\mathrm{sim}(\boldsymbol{b},\boldsymbol{w}_{i}))/\tau_{1})},  
\end{equation}
where $\tau_{1}$ is the temperature parameter, $\mathrm{sim}$ represents the cosine similarity,   $\boldsymbol{b}$ and $\boldsymbol{w}_{c}$ are the image and phrase features generated by CLIP, respectively, and $M$ is the size of the vocabulary. Then the image having the maximum similarity with the phrase is selected as the image sub-prompt.

For obtaining the text sub-prompt, we use a simple nearest-verb-search scheme, that is, finding the nearest verb (which is in a pre-built verb vocabulary) just before a specific object/location word. As shown in Fig.~\ref{fig:prompt base construction}, for the word ``kitchen'', the verb ``walk'' is located and then the phrase ``walk through the kitchen'' is extracted as the text sub-prompt. Finally, the image and text sub-prompts with the same visual object/location and action are formed as an aligned action prompt.




\subsection{Training and Inference}
\label{sec:Training}

\textbf{Modality Alignment Loss.} 
While an action prompt has the matched image and text sub-prompts, they may not be aligned in the feature space. To address this problem,  
following the contrastive learning paradigm used in CLIP \cite{radford2021learning} that enforces paired image and text features to be similar and non-paired ones to be distant, we use the infoNCE loss \cite{chen2020a} to encourage the feature alignment of the image and text sub-prompts in each action prompt:
\begin{equation}
\mathcal{L}_{a}=-\mathrm{log}(\frac{\mathrm{e}^{\mathrm{sim}(\mathbf{P}_{n}^{i}, \mathbf{P}_{n}^{u})/\tau_{2}}}{\mathrm{e}^{\mathrm{sim}(\mathbf{P}_{n}^{i}, \mathbf{P}_{n}^{u})/\tau_{2}}+\sum\limits_{\mathbf{\overline{P}}^{u}_{n}}\mathrm{e}^{\mathrm{sim}(\mathbf{P}_{n}^{i}, \mathbf{\overline{P}}^{u}_{n})/\tau_{2}}}),
\end{equation}
where $\tau_{2}$ is the temperature parameter,  $\mathbf{P}_{n}^{i}$ and $\mathbf{P}_{n}^{u}$ represent the features of the paired image and text sub-prompts of action prompt $p_{n}$, and $\mathbf{P}_{n}^{i}$ and $\mathbf{\overline{P}}^{u}_{n}$ denote the non-paired sub-prompts. Through the modality alignment loss, the action prompts can become more discriminative for guiding the learning of action-level modality alignment.

\textbf{Sequential Consistency Loss.}
Since an instruction usually refers to  different visual landmarks sequentially, the action prompts in the retrieved action prompt  set $\{p_{n}\}$ are also related to  different objects/locations. To encourage the agent to focus on related action prompts in the retrieved prompt set sequentially according to its visual observations, we develop a sequential consistency loss which is the sum of two single-modal consistency losses. Take the text modality as an example, at each timestep $t$, the attended text sub-prompt feature $\tilde{\mathbf{P}}_{t}^{u}$ and the attended instruction feature $\tilde{\mathbf{X}}_{t}$ are enforced to be close:
\begin{equation}
\mathcal{L}_{c}^{u} = ||\tilde{\mathbf{P}}_{t}^{u}-\tilde{\mathbf{X}}_{t}||^{2}.
\end{equation}
Similarly, define  $\mathcal{L}_{c}^{i} =||\tilde{\mathbf{P}}_{t}^{i}-\tilde{\mathbf{V}}_{t}||^{2}$, which is used to promote the similarity between the attended image sub-prompt feature $\tilde{\mathbf{P}}_{t}^{i}$ and the attended visual feature  $\tilde{\mathbf{V}}_{t}$. Then the sequential consistency loss $\mathcal{L}_{c}$ is obtained by:
\begin{equation}
\mathcal{L}_{c} = \mathcal{L}_{c}^{i} + \mathcal{L}_{c}^{u}.
\end{equation}
\textbf{Total Objective.}
Following most of existing works \cite{tan2019learning, hong2021vln, hong2020language}, we also use the navigation loss $\mathcal{L}_{n}$, which is the sum of an imitation loss  $\mathcal{L}_{IL}$ and a reinforcement learning loss $\mathcal{L}_{RL}$. Thus, the total training objective of our ADAPT is:
\vspace{-0.2cm}
\begin{equation}
\mathcal{L} = \mathcal{L}_{RL} + \lambda_{1}\mathcal{L}_{IL} + \lambda_{2}\mathcal{L}_{c} + \lambda_{3}\mathcal{L}_{a},
\end{equation}
where $\lambda_{1}$, $\lambda_{2}$, and $\lambda_{3}$ are the loss weights to balance the loss items.

\textbf{Inference.} During inference, the agent retrieves instruction-related action prompts from the action prompt base built in the training stage.

\begin{table*}[t]
    \centering
    \fontsize{6}{6}\selectfont
    \caption{Comparison with the SOTA methods on RxR. $^*$ indicates that the results are obtained by our re-implementation of the model.}
    \vspace{-1 em}
    \label{tab:com with sota on RxR}
    \resizebox{1.0\textwidth}{!}{
    {\renewcommand{\arraystretch}{1.2}
    \begin{tabular}{l|l||ccccc|ccccc}
    \specialrule{.1em}{.05em}{.05em}
        \multirow{2}{*}{Method}
        &\multirow{2}{*}{Model}
        & \multicolumn{5}{c|}{RxR Val Seen}
        & \multicolumn{5}{c}{RxR Val Unseen} \\
    \cline{3-12}
         && SR$\uparrow$ & SPL$\uparrow$ & CLS$\uparrow$ & nDTW$\uparrow$ & SDTW$\uparrow$  & SR$\uparrow$ & SPL$\uparrow$ & CLS$\uparrow$ & nDTW$\uparrow$ & SDTW$\uparrow$ \\
    \hline
       EnvDrop \cite{tan2019learning}&\multirow{4}{*}{ResNet-152}&48.1&44&61&57&40&38.5&34&54&51&32\\
       Syntax \cite{li2021improving} &&48.1&44&61&58&40&39.2&35&56&52&32\\

        VLN$\circlearrowright$BERT$^*$ 
        \cite{hong2021vln} &&50.9&45.4&60.3&56.9&41.3&45.5&39.3&56.6&52.9&36.3\\
               ADAPT (ours)&&\textbf{52.7}&\textbf{47.0}&\textbf{61.3}&\textbf{58.5}&\textbf{42.9}&\textbf{46.7}&\textbf{40.3}&\textbf{56.6}&\textbf{53.6}&\textbf{37.3}\\
                \hline
                VLN$\circlearrowright$BERT$^*$  \cite{hong2021vln}
        &\multirow{2}{*}{CLIP} &48.6&43.4&58.8&55.7&39.8&45.7&39.5&56.0&52.8&36.7\\
        ADAPT (ours)&&\textbf{50.3}&\textbf{44.6}&\textbf{59.6}&\textbf{56.3}&\textbf{40.6}&\textbf{46.9}&\textbf{40.2}&\textbf{57.2}&\textbf{54.1}&\textbf{37.7}\\
    \specialrule{.1em}{.05em}{.05em}
    \end{tabular}}}
    \label{tab:rxr}
    \vspace{-0.2cm}
\end{table*}

\begin{table*}[!htb]
	\fontsize{6}{6}\selectfont

\caption{Comparison with the SOTA methods on R2R. $^*$ indicates that the results are obtained by our re-implementation of the model.}
	\vspace{-0.2cm}
	\label{tab:com with sota}
		\resizebox{1.0\linewidth}{!}{
	{\renewcommand{\arraystretch}{1.2}
		\begin{tabular}{c||c|c|c|c|c|c|c|c|c|c|c|c}
			\specialrule{.1em}{.05em}{.05em}
			\multirow{2}{*}{Method}&\multicolumn{4}{c|}{Val Seen }&\multicolumn{4}{c|}{Val Unseen}&\multicolumn{4}{c}{Test Unseen}\cr\cline{2-13}
			&TL&NE $\downarrow$&SR $\uparrow$&SPL $\uparrow$&TL&NE $\downarrow$&SR $\uparrow$&SPL $\uparrow$&TL&NE $\downarrow$&SR $\uparrow$&SPL $\uparrow$\cr
			\hline
			
        
            Seq2Seq \cite{anderson2018vision}&11.33&6.01&39&-&8.39&7.81&22&-&8.13&7.85&20&18\\
            Speaker-Follower \cite{fried2018speaker}&-&3.36&66&-&-&6.62&35&-&14.82&6.62&35&28\\
            
   
          EnvDropout \cite{tan2019learning}&11.00&3.99&62&59&10.70&5.22&52&48&11.66&5.23&51&47\\  
                
        PREVALENT~\cite{hao2020towards} & 10.32 & 3.67 & 69 & 65 & 10.19 & 4.71 & 58 & 53 & 10.51 & 5.30 & 54 & 51 \\
		 VLN$\circlearrowright$BERT \cite{hong2021vln}&11.13&2.90&72&68&12.01&3.93&63&57&12.35&4.09&63&57\\
		 ADAPT (ResNet-152)&10.97&\textbf{2.54}&\textbf{76}&\textbf{72}&12.21&\textbf{3.77}&\textbf{64}&\textbf{58}&12.99&\textbf{3.79}&\textbf{65}&\textbf{59}\\  
		   \hline
    VLN$\circlearrowright$BERT$^*$ (CLIP) &11.37&3.17&70&66&12.03&3.81&65&58&12.73&4.26&61&55 \\
    ADAPT (CLIP)&11.39&\textbf{2.70}&\textbf{74}&\textbf{69}&12.33&\textbf{3.66}&\textbf{66}&\textbf{59}&13.16&\textbf{4.11}&\textbf{63}&\textbf{57}\\       
		 \specialrule{.1em}{.05em}{.05em}

			\end{tabular}}}
	\vspace{-0.4cm}
\end{table*}

\begin{table}[!htb]
	\fontsize{8}{8}\selectfont
	\caption{Ablation study of ADAPT on R2R Val Unseen. ResNet-152 and CLIP represent using different visual features. ADAPT-1: using action prompts only; ADAPT-2: using action prompts with the modality alignment loss; ADAPT-3: using action prompts with the sequential consistency loss; ADAPT-Full: our full model. All models are trained for 100K iterations.}
	\vspace{-0.2cm}
	\label{tab:Ablation study of ACT in R2R}
			\resizebox{1.0\linewidth}{!}{
	{\renewcommand{\arraystretch}{1.2}\begin{tabular}{c||c|c|c|c|c|c}
			 \specialrule{.1em}{.05em}{.05em}
			\multirow{2}{*}{Method}&\multicolumn{3}{c|}{ResNet-152}&\multicolumn{3}{c}{CLIP}\cr\cline{2-7}
			&NE $\downarrow$ &SR $\uparrow$ &SPL&NE $\downarrow$ &SR $\uparrow$ &SPL  $\uparrow$\cr
			\hline
			
            
          Baseline&4.17&60.4&54.7&4.11&61.5&55.3\\
          ADAPT-1&4.19&60.5&55.2&3.90&61.6&56.0\\
          ADAPT-2&4.16&61.7&55.4&\textbf{3.78}&62.8&56.3\\ 
        ADAPT-3&4.07&60.7&56.1&4.05&61.9&56.6\\ ADAPT-Full&\textbf{4.07}&\textbf{62.5}&\textbf{56.1}&4.10&\textbf{63.1}&\textbf{57.2}\\ 
         
      	 \specialrule{.1em}{.05em}{.05em}
		\end{tabular}}}
		\vspace{-0.4cm}
\end{table}

\begin{table*}[t]
    \centering
    \caption{ Results of the baseline \cite{hong2021vln} and our ADAPT on R2R Val Unseen with fewer training data. $^*$ indicates that the results are obtained by our re-implementation of the model.}
    \vspace{-1 em}
    \resizebox{1.0\textwidth}{!}{
    {\renewcommand{\arraystretch}{1}
    \begin{tabular}{l||cc|cc|cc|cc|cc|cc|cc|cc}
    \specialrule{.1em}{.05em}{.05em}
        \multirow{3}{*}{Model}&\multicolumn{8}{c|}{Scan}&\multicolumn{8}{c}{Instance}\\\cline{2-17}
        & \multicolumn{2}{c|}{20\%}
        & \multicolumn{2}{c|}{40\%}
        & \multicolumn{2}{c|}{60\%}& \multicolumn{2}{c|}{80\%}& \multicolumn{2}{c|}{20\%}
        & \multicolumn{2}{c|}{40\%}
        & \multicolumn{2}{c|}{60\%}& \multicolumn{2}{c}{80\%} \\
    \cline{2-17}
          & SR$\uparrow$ & SPL$\uparrow$ & SR$\uparrow$ & SPL$\uparrow$ & SR$\uparrow$ & SPL$\uparrow$ & SR$\uparrow$ & SPL$\uparrow$ & SR$\uparrow$ & SPL$\uparrow$ & SR$\uparrow$ & SPL$\uparrow$ & SR$\uparrow$ & SPL$\uparrow$ & SR$\uparrow$ & SPL$\uparrow$  \\
    \hline
        VLN$\circlearrowright$BERT$^*$ \cite{hong2021vln} &50.8&44.0&53.7&48.1&\textbf{57.7}&51.7&57.4&53.1&51.3&47.0&55.8&49.7&57.1&52.1&57.9&52.7 \\
        ADAPT (ours) &\textbf{52.5}&\textbf{46.4}&\textbf{55.1}&\textbf{48.8}&57.2&\textbf{51.8}&\textbf{59.1}&\textbf{53.3}&\textbf{52.5}&\textbf{47.3}&\textbf{56.6}&\textbf{49.8}&\textbf{58.8}&\textbf{53.5}&\textbf{59.4}&\textbf{54.6} \\

    \specialrule{.1em}{.05em}{.05em}
    \end{tabular}}}
   \vspace{-0.2cm} \label{tab:semisupervise}
\end{table*}
	
\section{Experiments}
\subsection{Experimental Setup}
\textbf{Datasets.}
We evaluate ADAPT on two public benchmarks, i.e., R2R \cite{anderson2018vision} and RxR \cite{ku2020room}. R2R \cite{anderson2018vision} includes 10,800 panoramic views and 7,189 trajectories. 
Since the baseline \cite{hong2021vln} is pretrained on English language data, we test our ADAPT on the English subset of RxR (both en-IN and en-US), which
includes 26,464 path-instruction pairs for training and 4,551 pairs in the val-unseen split.

\textbf{Evaluation Metrics.}
We use four popular metrics \cite{anderson2018vision} for the  performance evaluation on R2R: 1) Trajectory Length (TL) calculates the average length of the trajectory, 2) Navigation Error (NE) is the distance between target viewpoint  and agent stopping position, 3) Success Rate (SR) calculates the success rate of reaching the goal, and 4) Success rate weighted by Path Length (SPL) makes the trade-off between SR and TL. Three more metrics are used for RxR following other works \cite{ku2020room, li2021improving}: Coverage weighted by Length Score (CLS) \cite{jain2019stay}, Normalized Dynamic Time Warping (nDTW) \cite{ilharco2019general}, and Success rate weighted normalized Dynamic Time Warping (SDTW) \cite{ilharco2019general}.

\textbf{Implementation Details.}
All experiments are conducted on an NVIDIA V100 GPU. Two kinds of image features are used, i.e., the features extracted from a ResNet-152 \cite{he2016deep} pretrained on Places365 \cite{zhou2018places} and the features extracted through the visual encoder of CLIP \cite{shen2021how}. The model is trained for 300K and 100K iterations for R2R and RxR, respectively. The max sizes of the action prompt set are 60 and 100 for R2R and RxR, respectively. The instance whose number of retrieved action prompts less than the max size is padded.   The values of $\lambda_{1}$, $\lambda_{2}$, and $\lambda_{3}$ are 0.2,  0.01, and 0.0001, respectively. 
The same augmented data in \cite{hong2021vln} is used for R2R for fair comparison.
\subsection{Quantitative Results}
\textbf{Comparison with the State-of-the-Arts (SOTAs).} Table~\ref{tab:com with sota on RxR} and Table~\ref{tab:com with sota} give the comparison between existing methods and our ADAPT. Table~\ref{tab:com with sota  on RxR} shows that ADAPT with ResNet-152 feature outperforms previous SOTA methods on RxR. Moreover, ADAPT significantly improves the performance of the baseline \cite{hong2021vln} with different visual features in both Val Seen and Val Unseen settings on RxR, showing that introducing explicit action prompts can effectively promote the agent navigation capability.
From Table \ref{tab:com with sota} we can see that ADAPT (ResNet-152) establishes new SOTA results on R2R. Moreover, from the results of  VLN$\circlearrowright$BERT$^*$ (CLIP) and ADAPT (CLIP) we can find that by introducing the CLIP visual feature, both models show a performance enhancement in Val Unseen while a performance drop in both Val Seen and Test Unseen.
However, ADAPT (CLIP) outperforms VLN$\circlearrowright$BERT$^*$ (CLIP) on all the metrics, showing the effectiveness of the proposed method.   

\textbf{Ablation Study.} Table~\ref{tab:Ablation study of ACT in R2R} presents the ablation study results of ADAPT. As shown in Table~\ref{tab:Ablation study of ACT in R2R}, explicitly introducing the action prompts can effectively improve the performance of the strong baseline model \cite{hong2021vln}. By comparing the results between ``ADAPT-1'' and ``ADAPT-2'' we can find that introducing the modality alignment loss can effectively enhance the navigation performance, demonstrating that the action prompts with good discriminative power are useful for learning better action-level modality alignment. Comparing the results between ``ADAPT-2'' and ``ADAPT-Full'', we can see that the introduction of the sequential consistency loss further improves the navigation performance, which shows that attending to related action prompts sequentially is helpful for making correct action decision.  

To verify the generalization ability of ADAPT when a small amount of training data is available, we set up two training settings: ``Scan'' and ``Instance''. 
``Scan'' means that extracting part of the training scans with all the instances for training. ``Instance'' means that extracting all the training scans but with part of the instances for training. 
From the evaluation results given in Table~\ref{tab:semisupervise}, we can find that in both ``Scan'' and ``Instance'' settings, our ADAPT is superior over the strong baseline method, showing that by learning explicit action knowledge, the agent can have better generalization ability in different scenes. 

\begin{figure*}[t]
\begin{centering}
\includegraphics[width=0.95\linewidth]{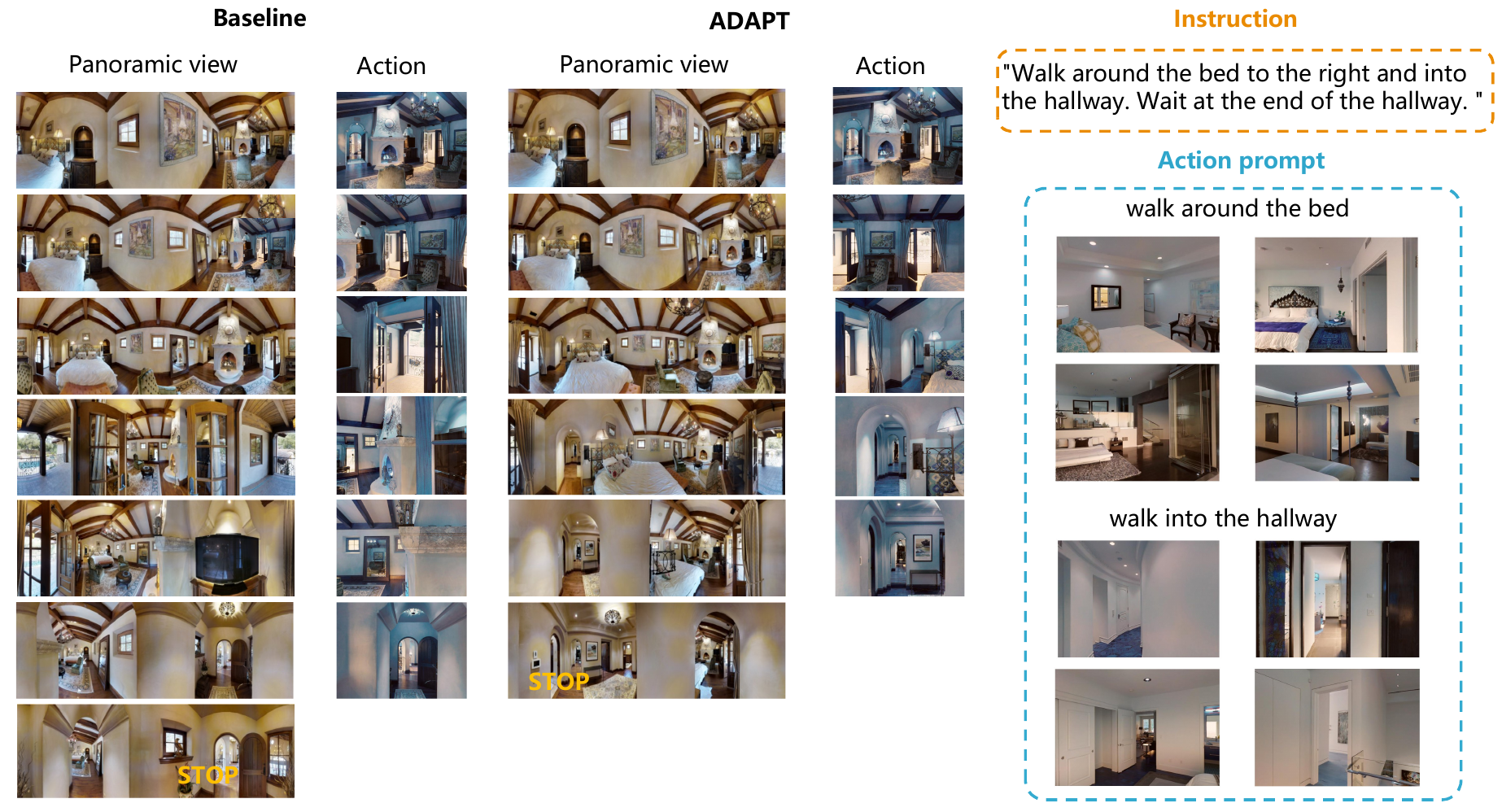}
\par\end{centering}
\caption{\label{fig:visualization}Visualization of panoramic views and action comparison in a trajectory example between the baseline  \cite{hong2021vln} and our ADAPT. 
}
\vspace{-0.4cm}
\end{figure*}

\begin{figure}[t]
\begin{centering}
\includegraphics[width=0.95\linewidth]{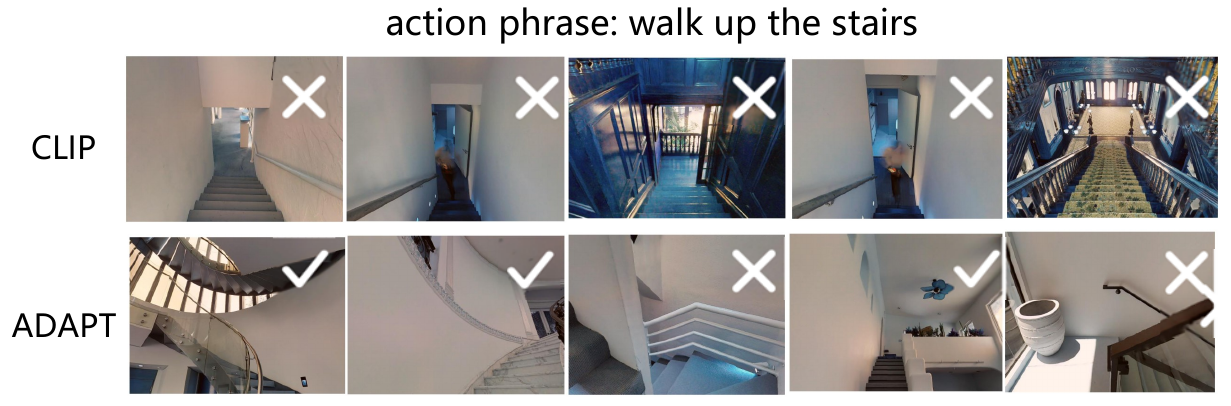}
\par\end{centering}
\caption{\label{fig:prompt alignment visualization}Action prompt alignment comparison between the CLIP features and the sub-prompt features of our ADAPT.
}
\vspace{-0.6cm}
\end{figure}

\subsection{Visualization}
We present some visualization results in this subsection to further analyze how introducing the explicit action prompts can contribute to correct navigation action decision. From Fig.~\ref{fig:visualization}  we can see that by introducing the action prompts related to ``walk around the bed'' and ``walk into the hallway'' in the instruction, our ADAPT can successful enforce the agent to choose the correct actions of walking around the bed and walking into the hallway in different visual observations. The baseline agent, however, leaves the original room and makes the wrong navigation trajectory. 

We further validate the action-level modality alignment ability of ADAPT by comparing the action prompt alignment between the CLIP
features and the sub-prompt features of ADAPT. For  the action phrase feature, the top 5  similar image features are retrieved from the object-related image  set. From Fig.~\ref{fig:prompt alignment visualization} we can find that compared with CLIP, ADAPT can perform better action-level modality alignment. Given the action phrase of ``walk up the stairs'', the top 5 results retrieved by CLIP from a set of stairs images all indicate the action of ``walk down'' the stairs. Our ADAPT, however, can obtain 3 images indicating the action of ``walk up'' the stairs in the top 5 results. 

\section{Conclusion and Limitation}
In this work, we propose modality-aligned action prompts (ADAPT), which prompts the VLN agent with explicit cross-modal action knowledge for enhancing the navigation performance. During navigation, the agent retrieves the action prompts from a pre-built action prompt base. Then the prompt-based instruction features are obtained for improving action decision. The CLIP model is used to collect high-quality action prompts into the prompt base. We also propose a modality alignment loss and a sequential consistency loss for training. Experiments on the public VLN benchmarks show the effectiveness of our ADAPT, which establishes new SOTA results. 
We hope this work can offer new directions for prompt-based navigation research.

With regards to the limitation of our work,  our constructed action prompt base in ADAPT contains more or less noise due to the ability of CLIP, the scene complexity and instruction diversity in the VLN task. The future work includes finding action prompts of better quality.

\section{Acknowledgement}
This work was supported in part by National Key R\&D Program of China under Grant No. 2020AAA0109700, National Natural Science Foundation of China (NSFC) No.61976233, Guangdong Province Basic and Applied Basic Research (Regional Joint Fund-Key) Grant No.2019B1515120039, Guangdong Outstanding Youth Fund (Grant No. 2021B1515020061), Shenzhen Fundamental Research Program (Project No. RCYX20200714114642083, No. JCYJ20190807154211365), National Natural Science Foundation of China under Grant No.62006255.  We thank MindSpore for the partial support
of this work, which is a new deep learning computing
framework\footnote{https://www.mindspore.cn/}.

\clearpage

{\small
\bibliographystyle{ieee_fullname}

}

\appendix

\section{Appendix}
In these supplementary materials, we first give more model details and  implementation details in Sec.~\ref{Model Details} and  Sec.~\ref{Implementation Details}, respectively. Then we present more quantitative results in Sec.~\ref{More Quantitative Results}.  
More visualization results are given in Sec.~\ref{Visualization}.

\subsection{Model Details}
\label{Model Details}
The self-attention module $\mathrm{SelfAttn}(\cdot)$ and the cross-modal attention module $\mathrm{CrossAttn}(\cdot)$ are the conventional multi-head self-attention modules in the standard Transformer. Specifically, given the query $Q$, the key $K$, and the value $V$, the self-attention of the $k$-th attention head at the $l$-th layer is calculated as follows:
\begin{equation}
H_{l,k} = \mathrm{Softmax}(\frac{QK^{\mathrm{T}}}{\sqrt{d_{h}}})V,
\end{equation}
where $d_{h}$ is the hidden dimension of the network. The attention $H_{l,k}$ is used to update the query $Q$ through the Feed-Forward Networks (FFN). In $\mathrm{SelfAttn}(\cdot)$, $Q$, $K$ and $V$ are obtained from the same single-modal feature. While in $\mathrm{CrossAttn}(\cdot)$, they are derived from the features of different modalities. The outputs of both modules are the self-attention values and the updated query features.

\subsection{Implementation Details}
\label{Implementation Details}
\textbf{Training.} Following~\cite{hong2021vln}, we use a batch size of 16 during training for both R2R and RxR. We employ a two-stage training strategy for ADAPT, i.e., first train the baseline model \cite{hong2021vln} until the performance is converged in Val Unseen, and then pick the model with the highest SPL from the first stage and continue to train it with our ADAPT. The learning rates in the first and second stages are set to 1e-5 and 1e-6, respectively. The optimizer is AdamW~\cite{loshchilov2018decoupled}. The sequential consistency loss is used in both the imitation learning training and the reinforcement learning training, while the modality alignment loss is used  only in the imitation learning training. During modality alignment learning, the non-paired sub-prompts for a specific sub-prompt are the sub-prompts in other samples in the same batch.

\textbf{Action Prompts.} Before feeding to the prompt encoder, the image and text sub-prompt features are extracted in advance through the visual and text encoders of CLIP \cite{radford2021learning}, respectively.  For accelerating the ADAPT training and inference, we perform the action prompt retrieval for each instruction in advance. The cosine similarity between two phrase features is calculated as the sentence similarity. The visual object/location vocabulary and the verb vocabulary are manually built by filtering the vocabulary of R2R given in \cite{anderson2018vision}. To mitigate the multiple object noises existing in collecting the action prompts,  we create a shared prompt base for R2R and RxR using the Fine-grained R2R\footnote{https://github.com/YicongHong/Fine-Grained-R2R}, which contains paired sub-paths and sub-instructions always regarding single object/location. By selecting image sub-prompts from the sub-path rather than the whole path, our model can largely mitigate the noise. 

\begin{table}[t]
	\fontsize{8}{8}\selectfont
	
	\caption{More quantitative comparison between some variants and our ADAPT.}
			\resizebox{1.0\linewidth}{!}{
	{\renewcommand{\arraystretch}{1.2}\begin{tabular}{c||c|c|c|c|c|c}
			 \specialrule{.1em}{.05em}{.05em}
		 \multirow{2}{*}{Method}&\multicolumn{3}{c|}{ResNet-152}&\multicolumn{3}{c}{CLIP}\cr\cline{2-7}
			&NE $\downarrow$ &SR $\uparrow$ &SPL&NE $\downarrow$ &SR $\uparrow$ &SPL  $\uparrow$\cr
			\hline

          Object &4.09&59.9&54.9&\textbf{4.01}&61.9&55.4\\ 
           
          Extra &4.41&59.0&54.2&4.29&60.7&55.1\\ 
     Whole  &\textbf{4.05}&61.8&55.3&4.14&60.5&55.2\\   ADAPT &4.07&\textbf{62.5}&\textbf{56.1}&4.10&\textbf{63.1}&\textbf{57.2}\\   
      		
         \specialrule{.1em}{.05em}{.05em}
		\end{tabular}}}
\label{tab:more results}
	\vspace{-0.2cm}	
\end{table}

\begin{figure}[t]
\begin{centering}
\includegraphics[width=0.95\linewidth]{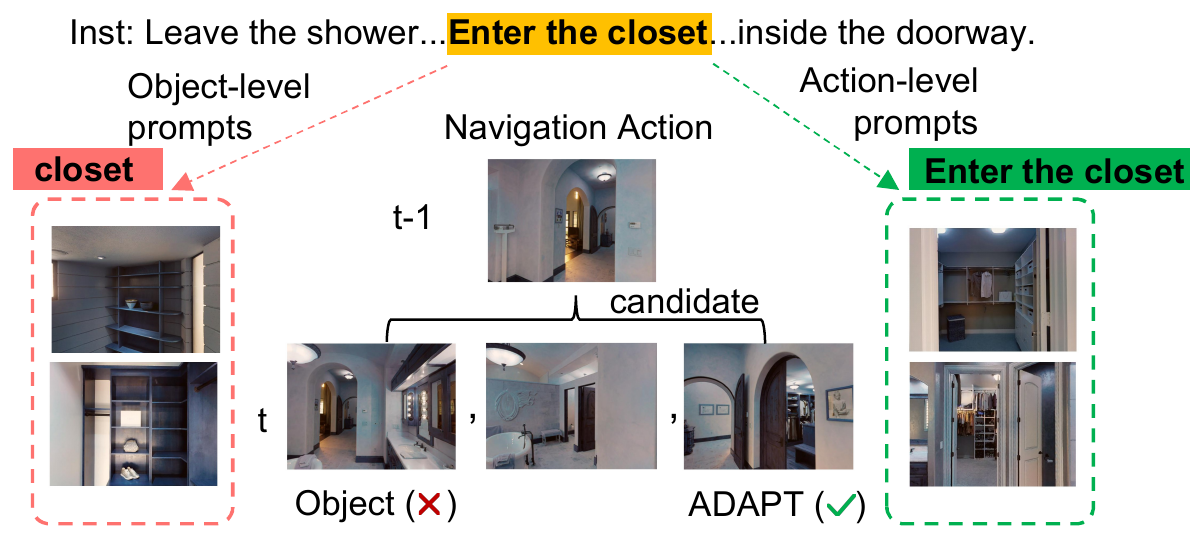}
\par\end{centering}

\caption{\label{fig:object}Action selection comparison between the models provided with object-level  prompts and action-level prompts. 
}
\vspace{-0.2cm}

\end{figure}

\begin{figure}[t]
\begin{centering}
\includegraphics[width=0.95\linewidth]{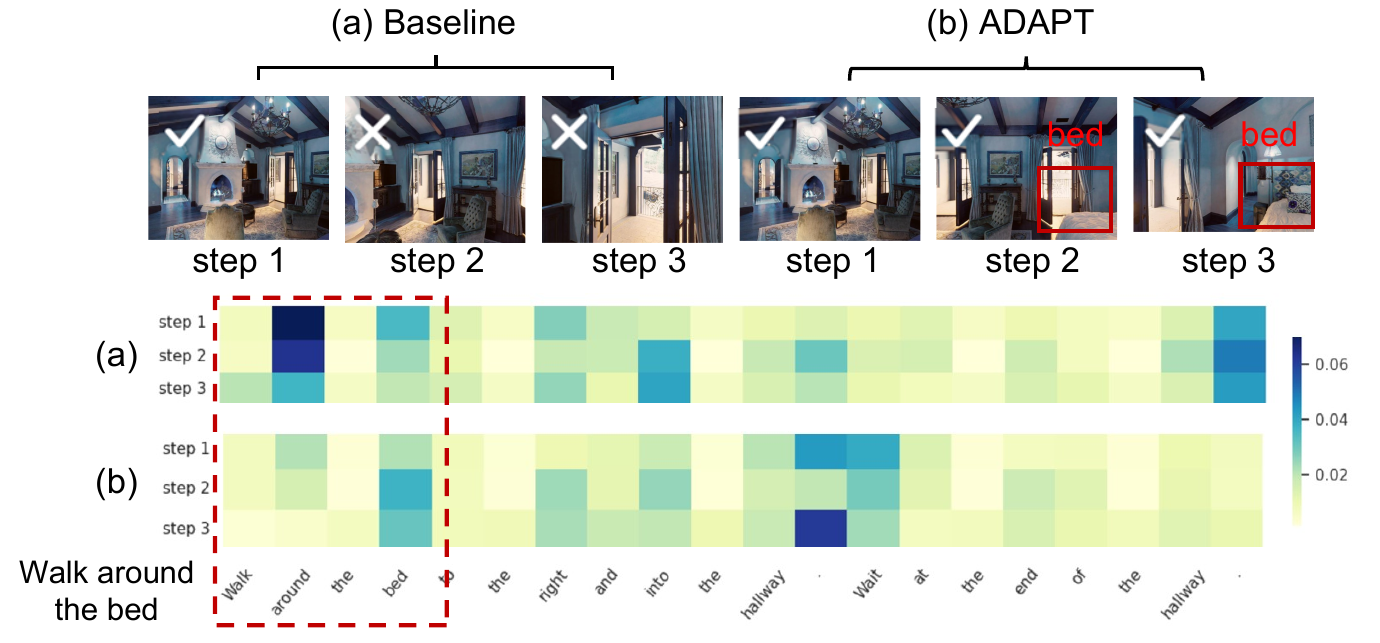}
\par\end{centering}

\caption{\label{fig:implicit}Language attention comparison between the baseline~\cite{hong2021vln} and ADAPT. 
}
\vspace{-0.2cm}

\end{figure}

\subsection{More Quantitative Results}
\label{More Quantitative Results}
In this subsection, we present the quantitative comparison between some variants, i.e., ``Object'', ``Extra'', ``Whole'' and our ADAPT in Table~\ref{tab:more results}. Specifically, ``Object'' means the paired
text and image sub-prompts are replaced by object words
and related object images, respectively. ``Extra'' means using action prompts as training samples directly. ``Whole'' means selecting image sub-prompts from the whole path rather than the sub-path in Fine-grained R2R. From the comparison results between ``Object'' and ADAPT we can observe that learning action-level alignment is more helpful for successful navigation than object-level alignment. The comparison results between ``Extra'' and ADAPT show the advantage of explicit action prompt
learning over implicit training. The comparison results between ``Whole'' and ADAPT demonstrates that ADAPT  effectively mitigates the multiple object noise during the action prompt collection.

\subsection{More Visualization Results}
\label{Visualization}
\textbf{Object-level Alignment vs. Action-level Alignment.} Fig.~\ref{fig:object} gives an action selection comparison between the models provided with object-level  prompts and action-level prompts. From Fig.~\ref{fig:object} we can observe that with action prompts indicating Enter the closet, ADAPT can successfully choose the action image which contains the closet door to enter it. The model provided with  object prompts about the closet, however, selects the wrong action. These results show the necessity of action-level alignment. 

\textbf{Implicit Learning vs. Explicit Learning.} We give a language attention  comparison between the baseline agent~\cite{hong2021vln} and ADAPT in Fig.~\ref{fig:implicit}, where we can find that both models  attend to the right instruction part in steps 1-3. However, the baseline conducts wrong modality alignment and action. This shows that compared with attention-based implicit  learning through simple navigation supervision, explicit action prompt learning  contributes to learning better modality alignment  for successful navigation.

\begin{figure*}[t]
\begin{centering}
\includegraphics[width=0.95\linewidth]{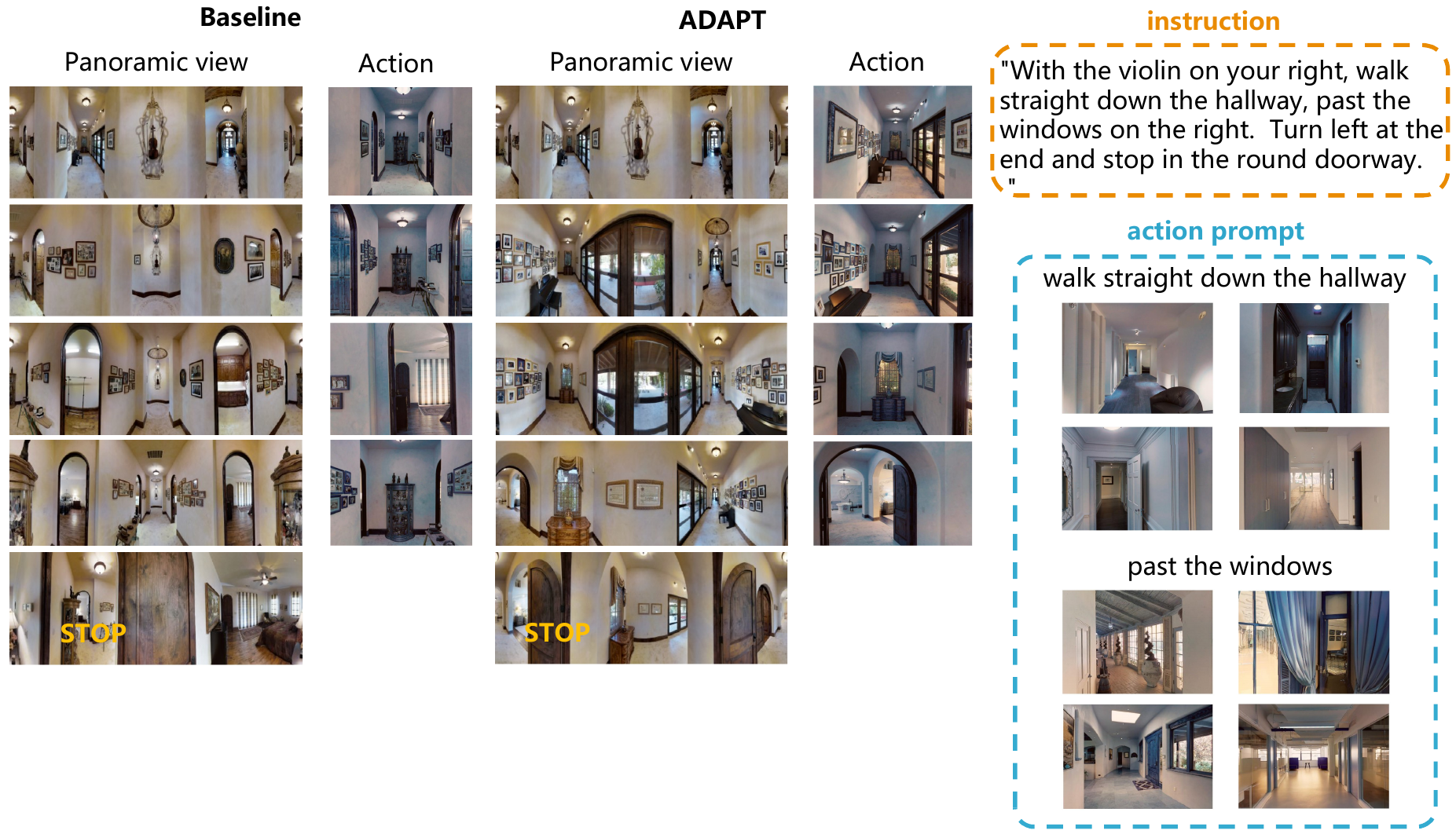}
\par\end{centering}
\caption{\label{fig:trajectory}Visualization of panoramic views and action comparison in a trajectory example between the baseline  \cite{hong2021vln} and our ADAPT.
}
\vspace{-0.4cm}
\end{figure*}

\begin{figure*}[t]
\begin{centering}
\includegraphics[width=0.95\linewidth]{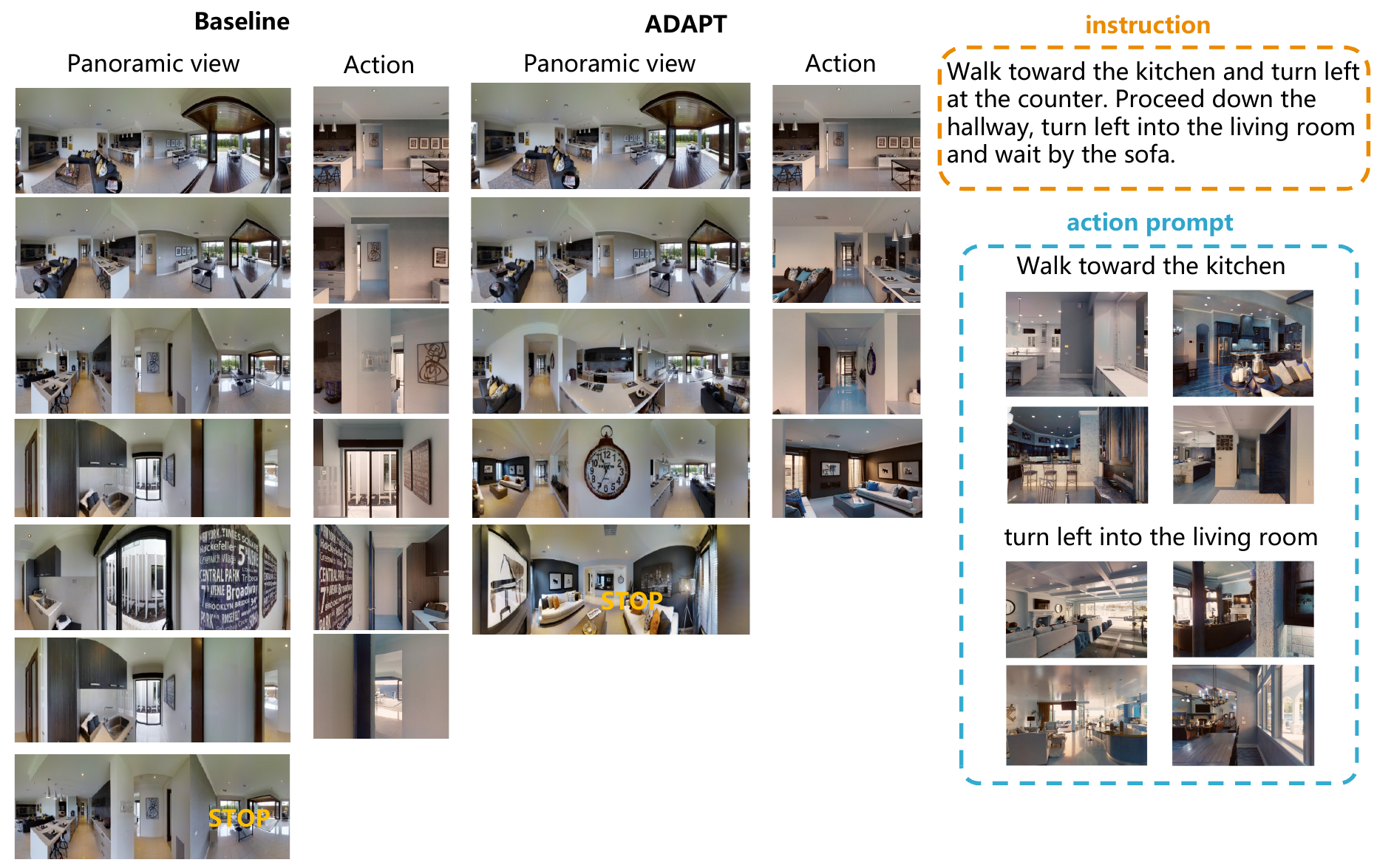}
\par\end{centering}
\caption{\label{fig:trajectory_2}Visualization of panoramic views and action comparison in a trajectory example between the baseline  \cite{hong2021vln} and our ADAPT.
}
\vspace{-0.4cm}
\end{figure*}

\textbf{Navigation Trajectories.} Fig.~\ref{fig:trajectory} and Fig.~\ref{fig:trajectory_2} present examples of the panoramic views and action comparison between the baseline~\cite{hong2021vln} and our ADAPT. From the visualization results we find that by introducing action prompts, ADAPT can make action decision accurately to accomplish successful navigation. For example, in Fig.~\ref{fig:trajectory}, with the help of the action prompts related to ``past the windows'', ADAPT makes the correct action of ``past the windows'' in the first two navigation steps. The baseline agent, however, fails to conduct the action of ``past the windows'' during navigation and thus makes a wrong trajectory.

\begin{figure*}[t]
\begin{centering}
\includegraphics[width=0.95\linewidth]{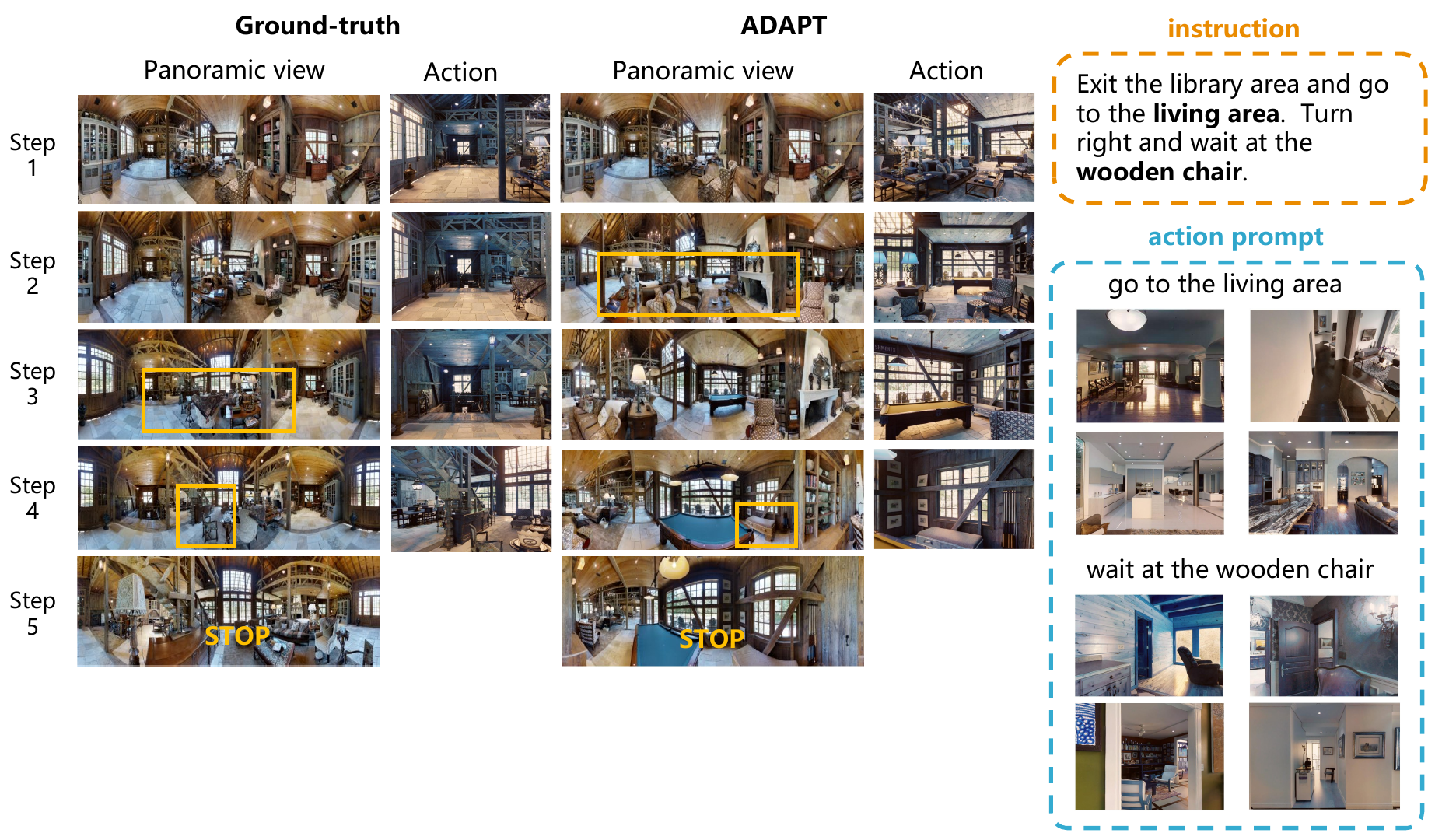}
\par\end{centering}
\caption{\label{fig:failure_case_1}A failed trajectory example of our ADAPT and the related  ground-truth. The yellow boxes indicate the instruction-related visual objects/locations appearing during the navigation trajectory.
}
\vspace{-0.4cm}
\end{figure*}

\begin{figure*}[t]
\begin{centering}
\includegraphics[width=0.95\linewidth]{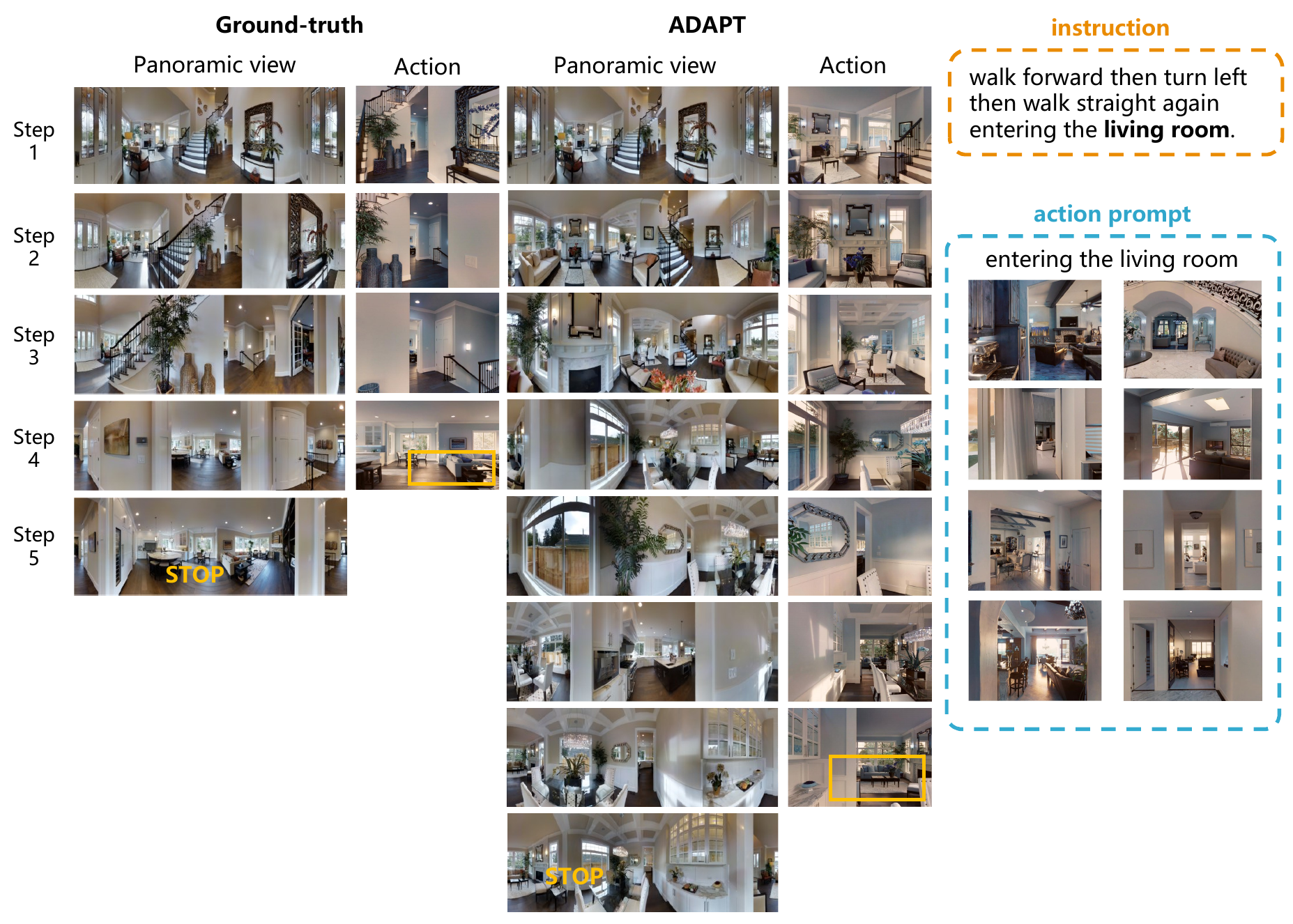}
\par\end{centering}
\caption{\label{fig:failure_case_2}A failed trajectory example of our ADAPT and the related  ground-truth. The yellow boxes indicate the instruction-related visual objects/locations appearing during the navigation trajectory.
}
\vspace{-0.4cm}
\end{figure*}

\textbf{Failure Cases of Navigation Trajectories.} Fig.~\ref{fig:failure_case_1} and Fig.~\ref{fig:failure_case_2} give failed trajectory examples of our ADAPT and the ground-truth. From the failure cases we can see that ADAPT may fail when the observations and instructions cause an ambiguity during navigation. From Fig.~\ref{fig:failure_case_1} we observe that although our ADAPT makes a wrong trajectory due to ambiguous observations of multiple ``living area'' and ``wooden chair'', it still conducts the correct actions of ``go to the living area'' and ``wait at the wooden chair'' in Step 2 and Step 4, respectively. From Fig.~\ref{fig:failure_case_2} we can find that due to the ambiguous instruction of ``walk forward and turn left'' without referring to concrete visual objects, our ADAPT makes the action of ``turn left'' before passing the stairs (the ground-truth one should be conducted after passing the stairs) and thus leads to a wrong trajectory. However, it still conducts the asked action of ``entering the living room'' at the end of the navigation. 

\begin{figure*}[t]
\begin{centering}
\includegraphics[width=0.95\linewidth]{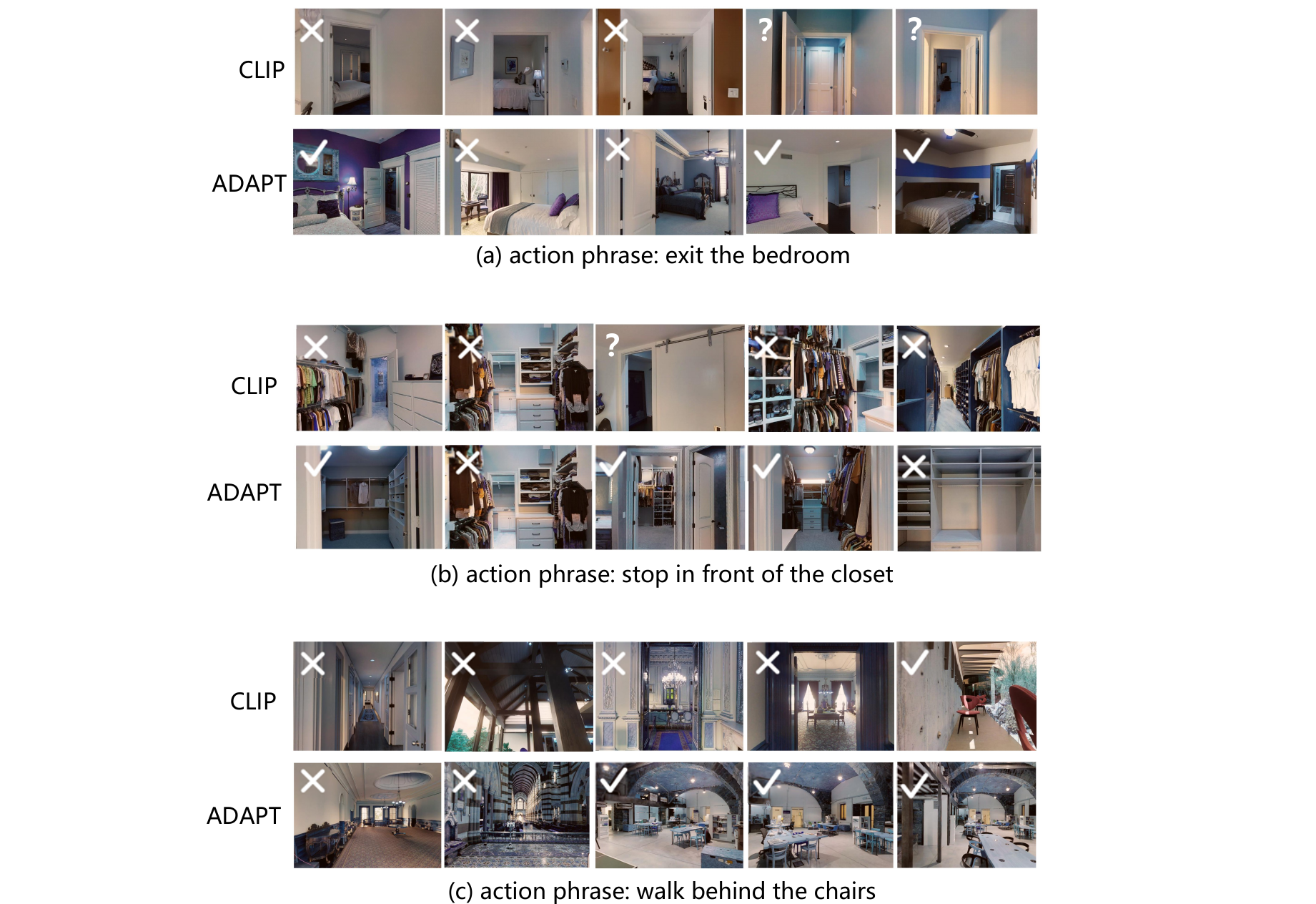}
\par\end{centering}
\caption{\label{fig:prompt alignment}Action prompt alignment comparison between the CLIP features and the sub-prompt features of our ADAPT. ``\checkmark'': correct; ``\xmark'': incorrect; ``\textbf{?}'': ambiguous.
}
\vspace{-0.4cm}
\end{figure*}

\textbf{Action Prompt Alignment.}
Fig.~\ref{fig:prompt alignment} presents additional results of action prompt alignment between the CLIP \cite{radford2021learning} features and the sub-prompt features of our ADAPT. For the action  phrase feature, the top 5 similar image features are retrieved from the object/location-related image sub-prompt set. From Fig.~\ref{fig:prompt alignment} we can observe that compared with CLIP, our ADAPT can perform better action-level modality alignment.
For example, in Fig.~\ref{fig:prompt alignment}(b) our ADAPT can effectively retrieve the closet images containing the appearances of the closet and its door through which the agent can make the action of ``stop in front of''.

\clearpage

\clearpage

\end{document}